\pdfoutput=1

\documentclass[11pt]{article}

\usepackage{ACL2023}
\usepackage{tikz}
\usepackage{colortbl}
\usepackage{array}
\usepackage{algorithm}
\usepackage{algorithmic}
\usepackage{graphicx}

\usepackage{colortbl}  
\usepackage{booktabs} 
\usepackage{times}
\usepackage{latexsym}
\usepackage{multirow}
\usepackage{enumerate}
\usepackage{enumitem}
\usepackage{amsmath}
\usepackage{amssymb}
\usepackage{xcolor}
\usepackage{tcolorbox}
\tcbuselibrary{most}
\usepackage{lipsum}

\usepackage[T1]{fontenc}

\usepackage[utf8]{inputenc}

\usepackage{microtype}

\usepackage{inconsolata}

\newcommand{\cellbg}[2][green!20]{%
  \begin{tikzpicture}[baseline=(text.base)]
    \node[fill=#1, inner sep=2pt, outer sep=0.6pt, anchor=base] (text) {#2};
  \end{tikzpicture}%
}
\usepackage{CJKutf8}

\NewDocumentCommand{\hy}
{ mO{} }{\textcolor{red}{\textsuperscript{\textit{Yue}}\textsf{\textbf{\small[#1]}}}}

\title{\texttt{1+1>2}: Can Large Language Models Serve as Cross-Lingual Knowledge Aggregators?}

\author{
\textbf{Yue Huang}$^{1}$\footnotemark[1], ~~\textbf{Chenrui Fan}$^{2}$\footnotemark[1], ~~\textbf{Yuan Li}$^{3}$, ~~\textbf{Siyuan Wu}$^{4}$, \\
\textbf{Tianyi Zhou}$^{2}$,  ~~\textbf{Xiangliang Zhang}$^{1}$\footnotemark[2], ~~\textbf{Lichao Sun}$^{5}$\\
\\
$^{1}$University of Notre Dame~ $^{2}$University of Maryland, College Park~ $^{3}$University of Cambridge\\ $^{4}$Huazhong University of Science and Technology $^{5}$Lehigh University \\
\texttt{\{yhuang37,xzhang33\}@nd.edu} ~~\texttt{cfan42@umd.edu}
}

\begin{document}
\begin{CJK}{UTF8}{gbsn}
\maketitle
\renewcommand{\thefootnote}{\fnsymbol{footnote}}
\footnotetext[1]{Equal contribution.}
\footnotetext[2]{Corresponding anthor.}

\begin{abstract}
Large Language Models (LLMs) have garnered significant attention due to their remarkable ability to process information across various languages. Despite their capabilities, they exhibit inconsistencies in handling identical queries in different languages, presenting challenges for further advancement. This paper introduces a method to enhance the multilingual performance of LLMs by aggregating knowledge from diverse languages. This approach incorporates a low-resource knowledge detector specific to a language, a language selection process, and mechanisms for answer replacement and integration. Our experiments demonstrate notable performance improvements, particularly in reducing language performance disparity. An ablation study confirms that each component of our method significantly contributes to these enhancements. This research highlights the inherent potential of LLMs to harmonize multilingual capabilities and offers valuable insights for further exploration.
\end{abstract}

\section{Introduction}

Large Language Models (LLMs) are increasingly recognized for their impressive capabilities in natural language processing (NLP). Employed across a variety of domains such as the medical sector \citep{liu2023deid, zhang2023biomedgpt}, software engineering \citep{qian2023communicative}, scientific research \citep{NEURIPS2023_bbb33018, li2024i}, and LLM-based agents \citep{liu2023agentbench, guo2024large, huang2024metatool, chen2024guiworld}, LLMs have demonstrated significant utility. Additionally, recent advancements in LLMs have expanded research \citep{qin2024multilingual, li2024xinstruction, xu2024survey, chen2024orion14b}, which focuses on enhancing their ability to process multiple languages and thereby increasing their accessibility and relevance across diverse linguistic demographics.

\begin{figure}
    \centering
    \includegraphics[width=1\linewidth]{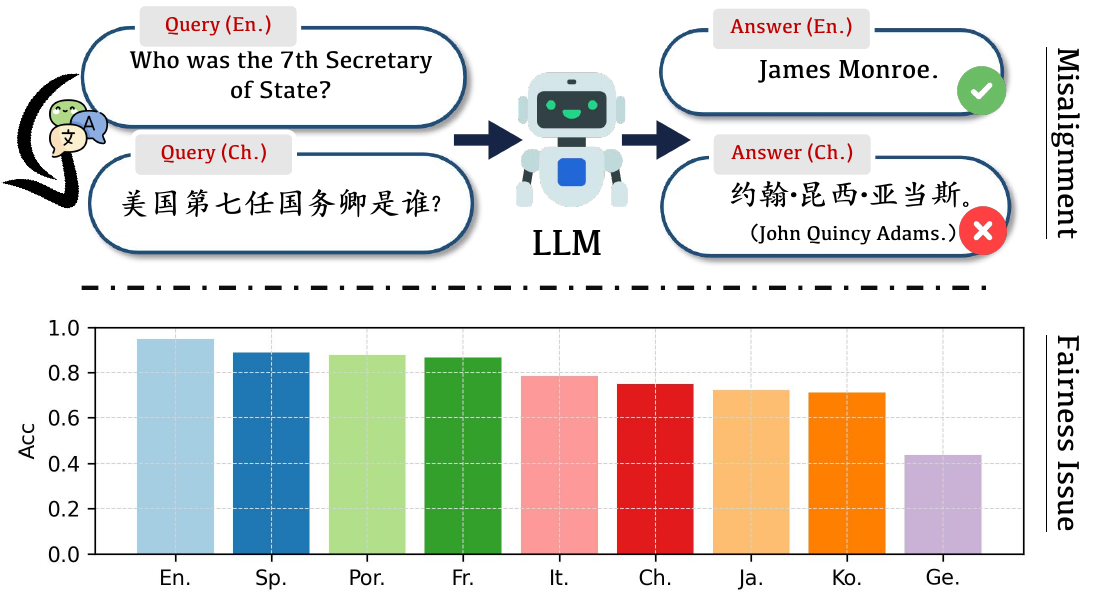}
    \caption{The top is an example of distinct answers to the same questions in different languages. The bottom is the GPT-4's performance on 300 queries in HalluEval \cite{li2023halueval} of nine different languages.}
    \label{fig:intro_fig}
\end{figure}

Despite these advancements, LLMs demonstrate inconsistencies when processing queries in different languages with the same meaning \cite{li2024quantifying}, as evidenced by the results in \autoref{fig:intro_fig}. This inconsistency not only diminishes the efficacy and fairness of LLMs but also signals underlying knowledge conflicts \citep{xu2024knowledge} that prevent these models from achieving true intelligence \citep{liu2023agentbench, huang2024metatool}. Furthermore, such inconsistency can erode trust in LLM applications, particularly when users from varied linguistic backgrounds cannot equally benefit from the technology \citep{li2023survey}.

To address the inconsistency problems in LLMs, we propose a novel method by leveraging the intrinsic capabilities of LLMs through integrating knowledge across different languages. Our approach begins with the development of a low-resource knowledge detector. This detector assesses whether a user's query involves knowledge that is underrepresented in the specific language. When the query does not feature low-resource knowledge, it is directly addressed by the LLMs. In contrast, if low-resource knowledge is detected, the LLMs will be required to select the most relevant target language to handle this specific knowledge. Once the target language is selected, the query is translated into this language, and the LLMs generate a response based on the translated query. This response either replaces the original answer or is integrated with it. Finally, the response is translated back to the original language of the query and delivered to the user. 

We conducted comprehensive experiments using six popular LLMs and five bilingual datasets (specifically, English and Chinese) to evaluate the effectiveness of our proposed method. The experimental results demonstrate that our approach successfully integrates knowledge from different languages, leading to the improvement in overall performance. Importantly, it significantly reduces the performance disparities across languages, thereby addressing the inconsistency issues inherent in LLMs and promoting fairness for downstream applications. Additionally, our ablation study confirms that both the low-resource knowledge detector and the language selection process are crucial to the improvements observed. Overall, our contributions are as follows:

\begin{itemize}[nolistsep, leftmargin=*]
    \item We posed an important challenge on the inconsistency of LLMs in downstream tasks, and the low-resource knowledge in a specific language can be brought from another language.
    \item Based on the observation, we propose a method that utilizes the LLMs' internal capability to enhance its performance on datasets in different datasets through a low-resources knowledge detector, language selection process, and answer replacement \& integration.
    \item We conduct extensive experiments on six popular LLMs and five bilingual datasets. The results show that our proposed method effectively enhances the performance of LLMs by integrating knowledge from different languages and reduce the performance gap in different languages.
\end{itemize}

\section{Related Work}

\subsection{Multilingual LLMs}

There has been a surge in research and work on Multilingual Large Language Models (MLLMs) \cite{qin2024multilingual, li2024xinstruction, xu2024survey, chen2024orion14b, etxaniz2023multilingual}. For instance, the InternLM, proposed by \citet{team2023internlm}, is a multilingual language model that has demonstrated excellent performance on multiple Chinese benchmarks. Similarly, PolyLM \cite{wei2023polylm} is another LLM trained using curriculum learning, surpassing other open-source models in multilingual tasks. Besides the above multilingual LLMs, the popular LLMs also include the ChatGLM series developed by \citet{du2022glm} and \citet{zeng2022glm}, and Baichuan series \citet{yang2023baichuan}. To improve model performance on multilingual tasks, \citet{muennighoff2023crosslingual} and \citet{zhang2023bayling} focus on utilizing multilingual training data to fine-tune the parameters. 

In terms of evaluation, \citet{lai2023chatgpt} assessed ChatGPT's performance across 37 different languages.CulturaX \cite{nguyen2023culturax} is a multilingual dataset containing 6.3 trillion tokens across 167 languages, aimed at promoting the development of multilingual LLMs. Additionally, M3Exam \cite{zhang2023m3exam} introduces a dataset derived from real and official human exam questions, designed for evaluating LLMs in a multilingual, multimodal, and multilevel context. BUFFET consolidates 15 varied tasks across 54 languages into a sequence-to-sequence format, offering a standardized set of few-shot examples and instructions \cite{asai2023buffet}. 

\subsection{Factuality in LLMs}

One way to improve the factuality of LLMs is the utilization of knowledge graphs (KGs)\cite{sun2024headtotail}. For instance, \citet{aburasheed2024knowledge} uses knowledge graphs to learn explainable recommendations. \citet{yang2024facts} suggests improving LLMs through the development of knowledge graph-enhanced LLMs, which offers a method to boost the factual reasoning capabilities of LLMs. \cite{sun2024thinkongraph} utilizes the LLM as an agent to interact with and navigate through the KGs, identifying relevant entities and relationships, and conducting reasoning with the knowledge it gathers.

Another method to enhance the factual knowledge of LLMs is the utilization of prompt engineering. Previous studies propose various prompt methods such as Chain-of-Thoughts (CoT) \cite{wei2023chainofthought} and Tree-of-Thoughts (ToT) \cite{yao2023tree}. Moreover, some studies use knowledge injection to enhance the domain capability of LLMs \cite{huang2024fakegpt}.

\subsection{Hallucination Mitigation}

A significant challenge associated with LLMs is their tendency to generate seemingly plausible yet fabricated responses, a phenomenon known as hallucination \cite{sun2024trustllm}. To address this issue and prevent misinformation \cite{10.1145/3589335.3651509}, recent research has introduced various hallucination mitigation strategies \cite{tonmoy2024comprehensive}. For example, \citet{feng2024dont} leverage multi-LLM collaboration to decrease hallucinations in LLM outputs. Additionally, \citet{guan2024mitigating} have developed a novel framework called Knowledge Graph-based Retrofitting (KGR), which integrates LLMs with KGs to minimize factual hallucinations during reasoning. Similarly, \citet{manakul2023selfcheckgpt} propose SelfCheckGPT, a sampling method that verifies the accuracy of responses from black-box models without the need for an external database.

\section{Methodology}

\begin{figure}
    \centering
    \includegraphics[width=\linewidth]{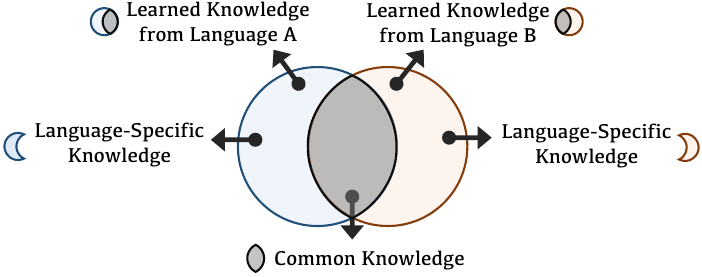}
    \caption{The knowledge domain of a multilingual LLM can be separated into multiple sections (the figure shows two). The language-specific knowledge (pure blue or pure orange) in one language can be utilized for improving the performance in other languages.}
    \label{fig:knowledge_domain}
\end{figure}

\begin{figure}
    \centering
    \includegraphics[width=\linewidth]{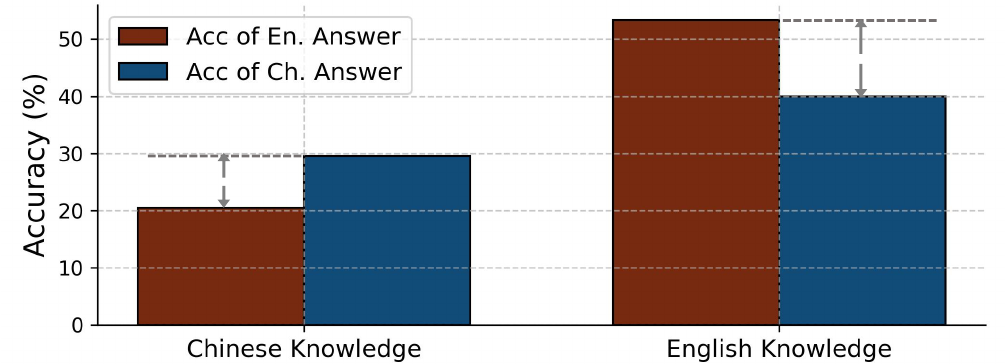}
    \caption{The average performance of six LLMs in five datasets. We show the accuracy of Chinese and English domain knowledge with the query/answer in Chinese and English.}
    \label{fig:motivation}
\end{figure}

\subsection{Motivation}

\begin{figure*}
    \centering
    \includegraphics[width=1\linewidth]{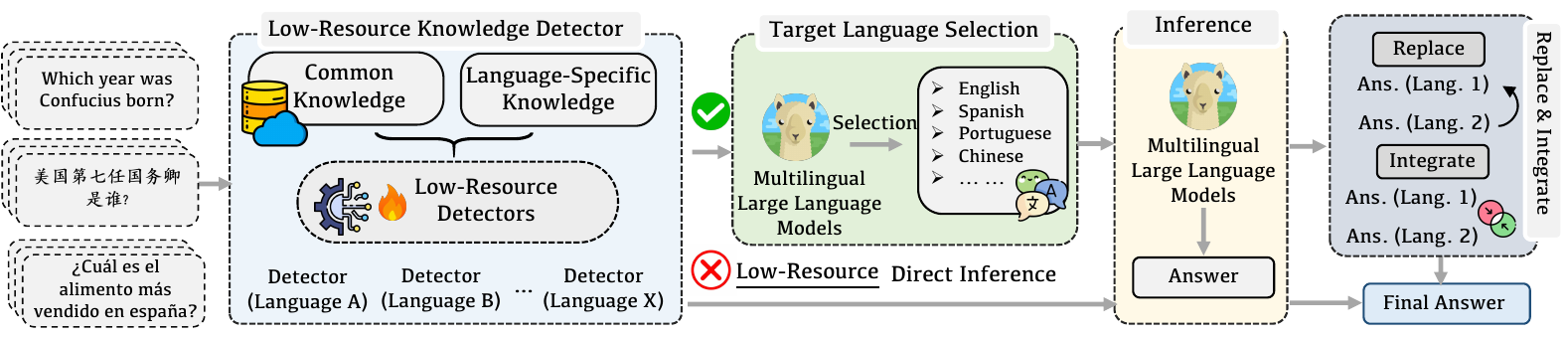}
    \caption{The proposed method begins with the query detection of low-resource knowledge powered by a detector. If low-resource knowledge is detected within the queries, LLMs then select the language most likely to yield the best answer. Answer replacement and integration are employed to formulate the final response.}
    \label{fig:pipeline}
\end{figure*}

Our proposed method draws inspiration from the distinct knowledge domains inherent to different languages. As illustrated in \autoref{fig:knowledge_domain}, language-specific knowledge can serve as supplementary information for another language. \autoref{fig:motivation} demonstrates that when queries related to English domain knowledge are posed in Chinese, the performance (i.e., accuracy) of LLMs declines compared to those posed in the English language. Furthermore, \autoref{fig:motivation2} reveals that LLMs often provide correct answers in only one of two languages for a given query, suggesting the potential to use the correct response to rectify inaccuracies in the other language. These observations underscore the potential to leverage the strengths of each language to enhance LLM performance across different languages. As shown in \autoref{fig:pipeline}, the proposed method includes three main modules: \textit{low-resource knowledge detection}, \textit{target language selection}, and \textit{answer replacement \& integration}.


\subsection{Construction of Low-Resource Dataset}
We first construct a low-resource dataset to measure current LLMs' multilingual knowledge transfer capacity. We also use this dataset to train our Low-Resource Knowledge Detector in section~\ref{sec:det}. We initial the dataset with the combination of various existing question-answering datasets including TriviaQA \cite{joshi2017triviaqa}, CMMLU \cite{li2024cmmlu}, HalluEval \cite{li2023halueval}, TruthfulQA \cite{lin2022truthfulqa}, MKQA \cite{longpre2021mkqa}, XQuAD \cite{Artetxe:etal:2019}, LC-QuAD \cite{trivedi2017lc}, KgCLUE \cite{xu-etal-2020-clue}. Moreover, we also construct a dataset that uses LLM-powered synthesized data to cover more knowledge and topics in the training corpus (We call it \textsc{MultiGen}). The details of the constructed dataset are shown in \autoref{app:constructed}.

To label these data items, we first use an LLM-Human collaboration to label the samples as Chinese-specific, English-specific, or common sense. Specifically, to confirm the correctness of the labeling, we infer the GPT-4 twice to label the samples with a temperature of 1.0 to enlarge the potential uncertainty of its output. We then conduct human inspections of the dataset where the labels are inconsistent in two labeling processes, to confirm the labeling and filter out the samples that are too hard or ambiguous for current LLMs. The statistics of the dataset can be found in Table~\ref{tab:dataset}.

\subsection{Low-Resource Knowledge Detector}
\label{sec:det}

The multilingual misalignment stems from the unbalanced training data as the knowledge with low data resources is less likely to be captured by the language model during the pretraining process. For example, queries about the details of Chinese history are not well answered by the model if asked in English as they appear less frequently in the English pretraining corpus. This phenomenon could be improved by fully utilizing the model's inherent capacity. To implement this process, we first adopt a low-resource knowledge detector to identify these low-resource queries and later borrow knowledge from other languages for help.

We train a classifier for each source language to identify the low-resource query for that language. This classifier separates the query about common sense and language-specified knowledge(e.g. Spanish query about Spanish culture) from the low-resources query(e.g. Spanish query about Turkish geography). Queries of the former class are fed into the normal pipeline of language generation while the latter queries are to be enhanced by the knowledge of other languages through our design of other modules. Given a query $x$ in the original language $L_o$, the low-resource knowledge detector $F_{L_o}$ works as follow:
\begin{equation}
    F_{L_o}(x)=\left\{
\begin{array}{rcl}
1     &,      & \text{$x$ is low-resource query of $L_o$}\\
0       &,     & \text{else}\\
\end{array} \right.
\end{equation}
We demonstrate in the experiment that a classifier is effective enough to distinguish low-resource queries from others. The construction of the training dataset of $F_L$ can be found in \autoref{sec:setup}.

The method is cost-effective as it does not require the translation of all queries to multiple languages considering that low-resource query is only a small part of user queries. The majority of user queries are related to common sense and knowledge specified in that language and do not need to go through the following process.

\subsection{Target Language Selection}
After selecting the low-resource query from the user's input, we later adopt a target language selection module to find the most suitable language for that question(e.g. translating a question in English about Chinese history into Chinese). Answering the query with its most resourceful language would improve output quality in terms of correctness and may offer more useful details to the user. We implement this process by prompting the LLM itself as the selection is model-dependent. Different LLMs may select different target languages due to their pretraining corpus. Given the prompt $P_{\text{sel}}$ to help select the target language, the low-resource query $x$, the procedure of Target Language Selection is defined as follows:
\begin{equation}
    x' \leftarrow \mathrm{Trans}\big(x, \text{LLM}(P_{\text{sel}}(x)) \big),
\end{equation}
where translator $\mathrm{Trans}(Q, L_t)$ translates the input $Q$ into target language $L_t$, and \text{LLM} is the large language model that selects the most suitable language for $x$ with prompt $P_{\text{sel}}$.

\subsection{Answer Replacement \& Integration}
After translating the original query $x$ to the query in target language $x'$, we use it to prompt the model for the answer in target language $a_t$. We simply translate the answer back to the original language to get the final answer $a_{\text{final}}$ for the user's understanding.
\begin{equation}
    a_{\text{final}} \leftarrow \mathrm{Trans}(a_t, L_o),
\end{equation}
where $L_o$ is the original language of the user's query. 

We also explore the integration of answers in the scenario of open-ended question answering (the prompt template is shown in \autoref{app:prompt_template}). We let the LLM combine and integrate the answer in the target language $a_t$ and the answer in the original language $a_o=\text{LLM}(x)$:
\begin{equation}
    a_{\text{final}} = \text{LLM}(P_{\text{int}}(a_t, a_o)),
\end{equation}
where $P_{\text{int}}$ is the prompt to help LLM integrate between $a_t$ and $a_o$, and $a_{\text{final}}$ is the final answer.

\begin{algorithm}
\caption{Proposed Method}
\label{alg:overall}
\begin{algorithmic}[1]
\REQUIRE Query $x$ in original language $L_o$
\ENSURE Final answer $a_{\text{final}}$
\STATE \textbf{Low-Resource Knowledge Detection:}
\STATE Train classifier $F_{L_o}$ for language $L_o$
\STATE $isLowResource \leftarrow F_{L_o}(x)$
\IF{$isLowResource == 1$}
    \STATE \textbf{Target Language Selection:}
    \STATE Define prompt $P_{\text{sel}}$ for selecting target language
    \STATE $L_t \leftarrow \text{LLM}(P_{\text{sel}}(x))$
    \STATE $x' \leftarrow \mathrm{Trans}(x, L_t)$
    \STATE \textbf{Answer Generation:}
    \STATE $a_t \leftarrow \text{LLM}(x')$
    \STATE $a_o \leftarrow \mathrm{Trans}(a_t, L_o)$
    \STATE \textbf{Answer Integration:}
    \STATE Define prompt $P_{\text{int}}$ for integrating answers
    \STATE $a_{\text{final}} \leftarrow \text{LLM}(P_{\text{int}}(a_t, a_o))$
\ELSE
    \STATE $a_{\text{final}} \leftarrow \text{LLM}(x)$
\ENDIF
\RETURN $a_{\text{final}}$
\end{algorithmic}
\end{algorithm}

\section{Experiments}

\noindent We chose English and Chinese for our experiments primarily due to their broad applicability and the availability of resources. Firstly, most LLMs, particularly open-source ones like the ChatGLM series, perform significantly in English and Chinese. This trend highlights the advanced development and optimization of LLMs for these languages, making them ideal for rigorous testing. Secondly, major LLM benchmarks and datasets predominantly focus on these two languages. For instance, besides English benchmarks or datasets, benchmarks such as HalluQA and AlignBench are primarily designed around English and Chinese, providing a robust framework for evaluating our methods. Lastly, the linguistic features and data availability in English and Chinese ensure comprehensive evaluation and validation of our approaches and suggest that our findings could be extrapolated to other languages. This potential for cross-linguistic application supports the broader relevance and utility of our study, choosing English and Chinese as both strategic and impactful.

\subsection{Experiment Setup}
\label{sec:setup}

\noindent\textbf{Training Datasets for Detectors.} As we need to train the low-resource detector for each language, for the dataset in English (\emph{e.g.}, TriviaQA) or the dataset in Chinese (\emph{e.g.}, CMMLU, KgCLUE), we translate them to another language (\emph{i.e.}, Chinese or English) through translation API \footnote{\url{https://fanyi.youdao.com/}}.

\noindent\textbf{Detailed Setting.} To ensure the reproducibility of results, the temperature parameter for all LLMs is set to 0. For ChatGPT, GPT-4, and Qwen-turbo, we use the official API. For Yi-34b, we use the API from Replicate\footnote{\url{https://replicate.com/}}. For ChatGLM3 and Llama3-Chinese, we deploy them locally for inference with a V100 (40G).

\begin{table}[]
\centering
\small
\setlength{\tabcolsep}{5pt}

\scalebox{0.9}{
\begin{tabular}{l|c|c|c|c|c}
\toprule[1pt]
\textbf{Dataset}    & \textbf{Chinese} & \textbf{Common} & \textbf{English} & \textbf{Total} & \textbf{Lang.} \\
\midrule
\textbf{TriviaQA}   & 21               & 754             & 1040             & 1815   & En.         \\
\textbf{CMMLU}      & 1200             & 2162            & 2751             & 6113   & Ch.         \\
\textbf{HalluEval}  & 28               & 923             & 1033             & 1984   & En.       \\
\textbf{TruthfulQA} & 9                & 322             & 212              & 543    & En.        \\
\textbf{MKQA}       & 71               & 315             & 1114             & 1500   & En.       \\
\textbf{XQuAD}      & 72               & 610             & 503              & 1185   & En.        \\
\textbf{LC-QuAD}    & 2                & 640             & 345              & 987    & En.        \\
\textbf{KgCLUE}     & 1218             & 610             & 172              & 2000   & Ch.        \\
\textbf{\textsc{MultiGen}}  & 1095             & 1121            & 1083             & 3299   & En.        \\
\midrule
\textbf{Total}      & 3716             & 7457            & 8253             & 19426  &   /    \\
\bottomrule[1pt]
\end{tabular}}
\caption{Dataset statistics of the low-resource knowledge detector. "Lang." is the original language for the dataset.}
\label{tab:dataset}
\end{table}

\begin{figure}
    \centering
    \includegraphics[width=1\linewidth]{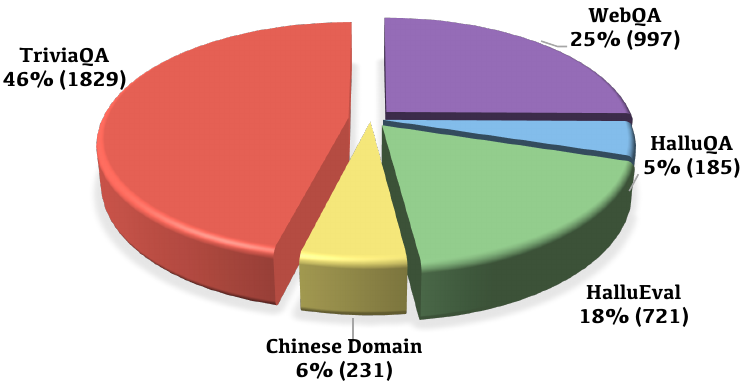}
    \caption{Statistics of the dataset in our experiments.}
    \label{fig:dataset_stat}
\end{figure}

\noindent\textbf{Test Datasets.} We selected five datasets for our study, comprising four pre-existing datasets and one that we developed in-house. The following criteria guided our selection:

\begin{itemize}[nolistsep, leftmargin=*]
    \item The datasets should not predominantly consist of common-sense questions (i.e., questions that are independent of linguistic background), as this minimizes the potential for LLMs to demonstrate improvement through linguistic knowledge.
    \item The datasets should maintain a balance in difficulty; they should not be overly simplistic or excessively challenging. Datasets that are too easy can lead to inflated performance metrics for LLMs, thereby reducing the potential for meaningful improvement. Conversely, datasets that are too challenging can degrade performance across all linguistic contexts, thereby constraining the opportunity to enhance performance in the target language by leveraging knowledge of additional languages.
\end{itemize}

\begin{table*}[!ht]
\centering
\small
\renewcommand\arraystretch{1.3}
\setlength{\tabcolsep}{3.2pt}

\scalebox{0.85}{\begin{tabular}{c|c|cc|cc|cc|cc|cc|cc}
\toprule[1pt]
\multicolumn{1}{c|}{}                                   &                                     & \multicolumn{2}{c|}{\textbf{ChatGLM3}}                             & \multicolumn{2}{c|}{\textbf{ChatGPT}}                              & \multicolumn{2}{c|}{\textbf{GPT-4}}                                & \multicolumn{2}{c|}{\textbf{Yi-34b}}                               & \multicolumn{2}{c|}{\textbf{Qwen-turbo}}     & \multicolumn{2}{c}{\textbf{Llama3-Ch.}}                      \\

 \cmidrule(lr){3-4}  \cmidrule(lr){5-6}  \cmidrule(lr){7-8}  \cmidrule(lr){9-10}  \cmidrule(lr){11-12} \cmidrule(lr){13-14}

\multicolumn{1}{c|}{\multirow{-2}{*}{\textbf{Dataset}}} & \multirow{-2}{*}{\textbf{Lang.}} & \textit{Origi.}           & \textit{Impro.}           & \textit{Origi.}           & \textit{Impro.}           & \textit{Origi.}           & \textit{Impro.}           & \textit{Origi.}           & \textit{Impro.}           & \textit{Origi.}           & \textit{Impro.}  & \textit{Origi.}           & \textit{Impro.}          \\
\midrule
                                                       & (en)                                & 18.03\%                         & 18.03\%                         & 57.98\%                         & 57.84\%                         & 67.13\%                         & 67.13\%                         & 42.86\%                         & 42.72\%                         & 29.31\%                         & 29.31\%      & 40.67\% & 40.67\%                   \\
\multirow{-2}{*}{\textbf{HalluEval}}                   & (ch)                                & \cellbg[green!20]{11.23\%} & \cellbg[green!20]{17.34\%} & \cellbg[green!20]{32.07\%} & \cellbg[green!20]{51.40\%} & \cellbg[green!20]{47.99\%} & \cellbg[green!20]{64.36\%} & \cellbg[green!20]{25.10\%} & \cellbg[green!20]{39.67\%} & \cellbg[green!20]{19.35\%} & \cellbg[green!20]{26.09\%} & \cellbg[green!20]{25.35\%} & \cellbg[green!20]{37.19\%} \\
\midrule
\midrule
                                                       & (en)                                & \cellbg[green!20]{20.00\%} & \cellbg[green!20]{25.95\%} & \cellbg[red!20]{34.27\%} & \cellbg[red!20]{30.90\%} & \cellbg[green!20]{51.89\%} & \cellbg[green!20]{54.05\%} & \cellbg[green!20]{38.38\%} & \cellbg[green!20]{47.03\%} & \cellbg[green!20]{25.97\%} & \cellbg[green!20]{37.57\%} & \cellbg[green!20]{22.83\%} & \cellbg[green!20]{19.57\%}\\
\multirow{-2}{*}{\textbf{HalluQA}}                     & (ch)                                & 22.16\%                         & 22.16\%                         & \cellbg[green!20]{21.91\%} & \cellbg[green!20]{24.16\%} & \cellbg[green!20]{49.73\%} & \cellbg[green!20]{51.35\%} & \cellbg[red!20]{45.95\%} & \cellbg[red!20]{44.86\%} & 43.65\%                         & 43.09\%     & \cellbg[green!20]{15.22\%} & \cellbg[green!20]{16.30\%}                    \\

\midrule
\midrule
                                          \textbf{Chinese}             & (en)                                & \cellbg[green!20]{9.52\%}  & \cellbg[green!20]{20.78\%} & 41.85\%                         & 42.73\%                         & \cellbg[green!20]{56.71\%} & \cellbg[green!20]{58.44\%} & \cellbg[green!20]{33.33\%} & \cellbg[green!20]{55.84\%} & \cellbg[green!20]{27.19\%} & \cellbg[green!20]{46.05\%} & \cellbg[red!20]{30.73\%} & \cellbg[red!20]{24.24\%} \\
\textbf{Domain}              & (ch)                                & 32.47\%                         & 32.47\%                         & 41.85\%                         & 41.85\%                         & 59.31\%                         & 59.74\%                         & 63.64\%                         & 63.20\%                         & 62.28\%                         & 61.84\%           & 18.61\%      & 18.61\%         \\

\midrule
\midrule
                                                       & (en)                                & 36.32\%                         & 36.32\%                         & 90.53\%                         & 90.37\%                         & 94.09\%                         & 94.09\%                         & 79.33\%                         & 79.17\%                         & 59.59\%                         & 59.47\%    & 77.27\% & 77.16\%                     \\
\multirow{-2}{*}{\textbf{triviaQA}}                    & (ch)                                & \cellbg[green!20]{21.33\%} & \cellbg[green!20]{31.95\%} & \cellbg[green!20]{54.60\%} & \cellbg[green!20]{82.67\%} & \cellbg[green!20]{82.77\%} & \cellbg[green!20]{91.90\%} & \cellbg[green!20]{59.43\%} & \cellbg[green!20]{75.56\%} & \cellbg[green!20]{41.53\%} & \cellbg[green!20]{52.99\%} & \cellbg[green!20]{43.92\%} & \cellbg[green!20]{65.17\%} \\
\midrule
\midrule
                                                       & (en)                                & \cellbg[green!20]{28.51\%} & \cellbg[green!20]{38.15\%} & 59.08\%                         & 58.88\%                         & \cellbg[green!20]{67.70\%} & \cellbg[green!20]{69.41\%} & \cellbg[green!20]{57.07\%} & \cellbg[green!20]{68.71\%} & \cellbg[green!20]{49.48\%} & \cellbg[green!20]{61.08\%} &
                                                       \cellbg[red!20]{50.00\%} & \cellbg[red!20]{48.09\%}
                                                       \\
\multirow{-2}{*}{\textbf{WebQA}}                       & (ch)                                & 48.69\%                         & 48.49\%                         & 57.35\%                         & 57.86\%                         & 72.52\%                         & 72.42\%                         & 76.93\%                         & 76.13\%                         & 71.12\%                         & 71.33\%     & \cellbg[green!20]{37.02\%} & \cellbg[green!20]{38.43\%}         \\          \bottomrule[1pt]
\end{tabular}}
\caption{Six LLMs' performance on our proposed method.}
\label{tab:proposed_performance}
\end{table*}

\noindent For all datasets in our study, we select QA-pair samples from them and do not use other extra data to facilitate our evaluation. Totally, we select five datasets for evaluating our method. These include four existing dataset: TriviaQA \cite{joshi2017triviaqa}, HaluEval \cite{li2023halueval}, HalluQA \cite{cheng2023evaluating}, and WebQA \cite{li2016dataset}. We show the statistics of the datasets we selected in \autoref{fig:dataset_stat} and the details are shown in \autoref{app:dataset_details}. In addition to the four datasets mentioned above, we have constructed a bilingual Chinese-English dataset tailored to the Chinese domain. Details of the construction process are provided in \autoref{app:construct_domain}.

\noindent \textbf{Models.} We carefully select six popular LLMs including proprietary and open-source LLMs that master both English and Chinese: ChatGPT \citep{ChatGPT}, GPT-4 \citep{GPT-4}, ChatGLM3 \citep{zeng2022glm, du2022glm}, Yi-34b \citep{ai2024yi}, Qwen-turbo \cite{qwen}, and LLama3-Chinese \citep{llama3-chinense}.

\subsection{Main Results} 
We evaluate the effectiveness of our proposed method on five benchmark datasets and six popular LLMs mentioned above. Each dataset is translated into a Chinese and an English version for later assessment. We first infer the models with the queries in the dataset to get the generated answers. We then leverage GPT-4 as the judge model to compare each generated answer with the reference answer in the dataset to see if the model produces a correct output. We calculate the generation accuracy and present the result in Table~\ref{tab:proposed_performance}.
We mark the result in \colorbox{green!20}{green} where there is a significant improvement of more than $1\%$ and mark the result in \colorbox{red!20}{red} if the accuracy decrease by more than $1\%$.

As can be seen from the table, our method can effectively improve the performance of the model in many scenarios. To be specific, the performance of the GPT-4 model on the HalluEval dataset in Chinese improves significantly from $47.99\%$ to $64.36\%$. This means there still exists a large cross-lingual knowledge gap in advanced models such as GPT-4 and our method successfully leverages the knowledge across languages to enhance the model's performance.
It is important to notice that the improvements do not rely on other models or online resources, they exist due to our leverage of the model's inherent capacity. 

\begin{figure}[]
    \centering    \includegraphics[width=\linewidth]{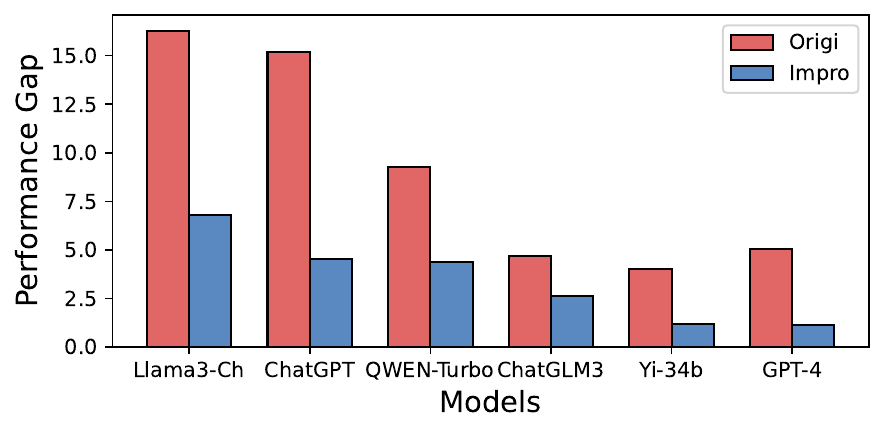}
    \caption{The average performance gap on datasets before and after applying our method.}
    \label{fig:gap}
\end{figure}

It can also be observed from Table~\ref{tab:proposed_performance} that most improvements happen in the language that is different from that of the original dataset, which is also the part where the models suffer from a weaker performance. The comparison of the cross-lingual performance gap before and after applying our method is shown in Figure~\ref{fig:gap}. The figure showcases that our method could significantly reduce the knowledge gap between languages in all LLMs we evaluate, thus improving the fairness of the application for users of different languages.

\subsection{Ablation Study}
As our generation pipeline consists of several parts, we conduct an ablation study to validate their effectiveness and expenses.

\begin{figure}[h]
    \centering    \includegraphics[width=0.8\linewidth]{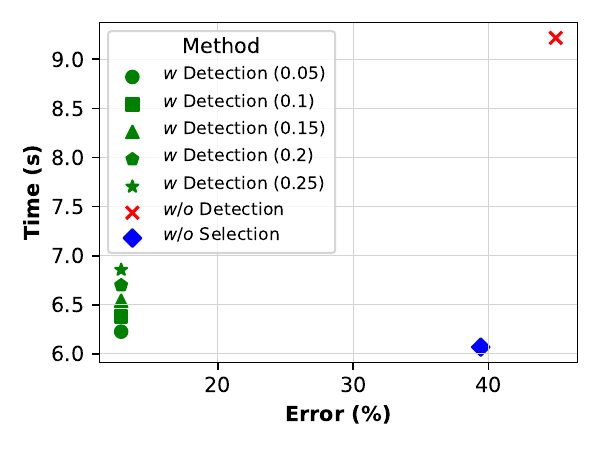}
    \caption{The relationship of time efficiency and error rate. The error rate is the percentage sum of all decreasing in five datasets (value in red on \autoref{tab:proposed_performance}, \autoref{tab:performance_without_detector} (\textit{w/o} Detection), \autoref{tab:performance_without_selection} (\textit{w/o} Selection)). }
    \label{fig:time_error}
\end{figure}

\noindent\textbf{The Impact of the Low-resource Detector.} The low-resource detector serves as a filter to sift the language-specific queries from the majority of the queries that involve only commonsense, thus improving efficiency and reducing the expense of the pipeline. As can be observed in Figure~\ref{fig:time_error}, a low-resource query detector would significantly reduce the average inference time per sample from more than 9 seconds to less than 6.5 seconds if the ratio of the low-resource queries is 0.05 in the dataset. When the ratio of the low-resource query in the dataset increases, the detector passes more samples into the translation pipeline and increases the average inference time.

Another intriguing finding is that the low-resource detector would increase the model performance. As shown in Table~\ref{tab:performance_without_detector}, the performance of the pipeline is unstable when we remove the low-resource detector. The overall performance would also drop as we observed in Figure~\ref{fig:time_error}. This indicates that the detector and LLM itself can be complementary. The full result of the models' performance without the low-resource detector can be found in Table~\ref{tab:performance_without_detector_full}.

\begin{table}[]
\centering
\small
\renewcommand\arraystretch{1.2}

\resizebox{\linewidth}{!}{\begin{tabular}{c|c|cc|cc|cc}
\toprule
\multicolumn{1}{c|}{}                                   &                                     & \multicolumn{2}{c|}{\textbf{Yi-34b}}                               & \multicolumn{2}{c|}{\textbf{Qwen-turbo}}     & \multicolumn{2}{c}{\textbf{Llama3-Ch.}}                      \\

 \cmidrule(lr){3-4}  \cmidrule(lr){5-6} \cmidrule(lr){7-8}

\multicolumn{1}{c|}{\multirow{-2}{*}{\textbf{Dataset}}} & \multirow{-2}{*}{\textbf{Lang.}} & \textit{Origi.}           & \textit{Impro.}           & \textit{Origi.}           & \textit{Impro.}  & \textit{Origi.}           & \textit{Impro.}          \\
\midrule

                                                       & (en)                                & \cellbg[red!20]{42.86\%} & \cellbg[red!20]{41.75\%} & 29.31\%                         & 29.59\%       & 40.67\% &        40.67\%         \\
\multirow{-2}{*}{\textbf{HalluEval}}                   & (ch)                                & \cellbg[green!20]{25.10\%} & \cellbg[green!20]{39.81\%} & \cellbg[green!20]{19.35\%} & \cellbg[green!20]{26.51\%} & \cellbg[green!20]{25.35\%} & \cellbg[green!20]{37.33\%} \\
\midrule
\midrule
                                                       & (en)                                & \cellbg[green!20]{38.38\%} & \cellbg[green!20]{47.03\%} & \cellbg[green!20]{25.97\%} & \cellbg[green!20]{37.57\%} & \cellbg[red!20]{22.83\%} & \cellbg[red!20]{18.48\%} \\
\multirow{-2}{*}{\textbf{HalluQA}}                     & (ch)                                & 45.95\%                         & 45.95\%                         & \cellbg[red!20]{43.65\%} & \cellbg[red!20]{39.78\%} & \cellbg[green!20]{15.22\%} & \cellbg[green!20]{20.65\%}  \\
\midrule
\midrule
                                       \textbf{Chinese}                & (en)                                & \cellbg[green!20]{33.33\%} & \cellbg[green!20]{57.58\%} & \cellbg[green!20]{27.19\%} & \cellbg[green!20]{48.25\%} & \cellbg[red!20]{30.74\%} & \cellbg[red!20]{24.24\%} \\
\textbf{ Domain}             & (ch)                                & \cellbg[red!20]{63.64\%} & \cellbg[red!20]{57.14\%} & 62.28\%                         & 62.28\%     & \cellbg[green!20]{18.61\% } & \cellbg[green!20]{22.51\%}                    \\
\bottomrule[1pt]
\end{tabular}}
\caption{Selected LLMs' performance on the setting without a low-resource detector.}
\label{tab:performance_without_detector}
\end{table}

\vspace{-0.3cm}

\begin{table}[]
\centering
\small
\renewcommand\arraystretch{1.2}

\resizebox{\linewidth}{!}{\begin{tabular}{c|c|cc|cc|cc}
\toprule[1pt]
\multicolumn{1}{c|}{}                                   &                                     & \multicolumn{2}{c|}{\textbf{Yi-34b}}                               & \multicolumn{2}{c|}{\textbf{Qwen-turbo}}     & \multicolumn{2}{c}{\textbf{Llama3-Ch.}}                      \\

 \cmidrule(lr){3-4}  \cmidrule(lr){5-6} \cmidrule(lr){7-8}

\multicolumn{1}{c|}{\multirow{-2}{*}{\textbf{Dataset}}} & \multirow{-2}{*}{\textbf{Lang.}} & \textit{Origi.}           & \textit{Impro.}           & \textit{Origi.}           & \textit{Impro.}  & \textit{Origi.}           & \textit{Impro.}          \\
\midrule
                                                       & (en)                                & {42.86\%} & {42.72\%} & 29.31\%                         & 29.17\%   & 40.67\% & 40.25\%                       \\
\multirow{-2}{*}{\textbf{HalluEval}}                   & (ch)                                & \cellbg[green!20]{25.10\%} & \cellbg[green!20]{42.58\%} & \cellbg[green!20]{19.35\%} & \cellbg[green!20]{28.61\%} & \cellbg[green!20]{25.34\%} & \cellbg[green!20]{39.97\%} \\
\midrule
\midrule
                                                       & {\color[HTML]{212121} (en)} & \cellbg[green!20]{38.38\%} & \cellbg[green!20]{46.49\%} & \cellbg[green!20]{25.97\%} & \cellbg[green!20]{38.67\%} & \cellbg[red!20]{22.83\%} & \cellbg[red!20]{21.74\%} \\
\multirow{-2}{*}{\textbf{HalluQA}}                     & (ch)                                & \cellbg[red!20]{45.95\%} & \cellbg[red!20]{44.86\%} & \cellbg[red!20]{43.65\%} & \cellbg[red!20]{41.44\%} & 15.21\% & 15.76\% \\
\midrule
\midrule
                                          \textbf{Chinese}             & (en)                                & \cellbg[green!20]{33.33\%} & \cellbg[green!20]{61.04\%} & \cellbg[green!20]{27.19\%} & \cellbg[green!20]{58.33\%} & 
                                                       \cellbg[red!20]{30.74\%} & \cellbg[red!20]{19.48\%}\\

\textbf{Domain}              & (ch)                                & \cellbg[red!20]{63.64\%} & \cellbg[red!20]{60.17\%} & \cellbg[red!20]{62.28\%} & \cellbg[red!20]{58.33\%}  & 18.61\% & 18.61\% \\
\bottomrule[1pt]
\end{tabular}}
\caption{Selected LLMs' performance on the setting without language selection.}
\label{tab:performance_without_selection}
\end{table}

\begin{table*}[t]
\centering
\small
\setlength{\tabcolsep}{3.5pt}

\begin{tabular}{c|c|cc|cc|cc|cc|cc|cc}
\toprule[1pt]
                                     &                                  & \multicolumn{2}{c|}{\textbf{ChatGPT}}                        & \multicolumn{2}{c|}{\textbf{GPT-4}}                          & \multicolumn{2}{c|}{\textbf{ChatGLM3}}                       & \multicolumn{2}{c|}{\textbf{Yi-34b}}                         & \multicolumn{2}{c}{\textbf{Qwen-Turbo}}  & \multicolumn{2}{c}{\textbf{Llama3-Ch.}}                  \\
                                     \cmidrule(lr){3-4} \cmidrule(lr){5-6} \cmidrule(lr){7-8} \cmidrule(lr){9-10} \cmidrule(lr){11-12} \cmidrule(lr){13-14}
\multirow{-2}{*}{\textbf{Type}}      & \multirow{-2}{*}{\textbf{Lang.}} & \textit{Origi.}        & \textit{Impro.}         & \textit{Origi.}        & \textit{Impro.}         & \textit{Origi.}        & \textit{Impro.}         & \textit{Origi.}        & \textit{Impro.}         & \textit{Origi.}        & \textit{Impro.}         & \textit{Origi.}        & \textit{Impro.}         \\
\midrule
                                     & \textbf{(ch)}                    & \cellbg[green!20]{4.98} & \cellbg[green!20]{5.26} & 6.90                          & 6.85                     & 4.15 & 4.09 & \cellbg[green!20]{5.82} & \cellbg[green!20]{5.91} & 5.88                         & 5.87      & \cellbg[red!20]{3.71} &   \cellbg[red!20]{3.58}                \\
\multirow{-2}{*}{\textbf{Integrate}}   & \textbf{(en)}                    & \cellbg[green!20]{5.92} & \cellbg[green!20]{6.02} & \cellbg[green!20]{7.32} & \cellbg[green!20]{7.54} & \cellbg[green!20]{4.02} & \cellbg[green!20]{4.07} & \cellbg[green!20]{5.86} & \cellbg[green!20]{6.13} & \cellbg[green!20]{5.59} & \cellbg[green!20]{5.78} & 4.60  & 4.60 \\
\midrule
\midrule
                                     & \textbf{(ch)}                    & \cellbg[green!20]{4.98} & \cellbg[green!20]{5.47} & 6.90  & 6.98 & {4.15} & {4.16} & \cellbg[green!20]{5.82} & \cellbg[green!20]{6.23} & \cellbg[green!20]{5.88} & \cellbg[green!20]{6.00}   & \cellbg[green!20]{3.71}  & \cellbg[green!20]{3.93}\\
\multirow{-2}{*}{\textbf{Replace}} & \textbf{(en)}                    & \cellbg[green!20]{5.92} & \cellbg[green!20]{5.97} & \cellbg[red!20]{7.32} & \cellbg[red!20]{7.12} & \cellbg[green!20]{4.02} & \cellbg[green!20]{4.26} & \cellbg[green!20]{5.86} & \cellbg[green!20]{6.25} & \cellbg[green!20]{5.59} & \cellbg[green!20]{5.88} & \cellbg[red!20]{4.60}  & \cellbg[red!20]{4.54} \\
\bottomrule[1pt]
\end{tabular}
\caption{Model performance on AlignBench \cite{liu2023alignbench} in the setting of answer replacement and integration.}
\label{tab:integration}
\end{table*}

\noindent\textbf{The Impact of the Language Selection Module.} The language selection module can choose the proper language to answer the question with model-specific choice. It is also flexible in the multi-lingual setting as the resulting target language can be more than two as we test. However, we still validate its effectiveness in the bi-lingual setting, comparing it with the strategy of using the opposite language when the query is detected as low-resources, and show our result in Table~\ref{tab:performance_without_selection}. The trade-off between its cost and error can also be found in Figure~\ref{fig:time_error}.

As we can see from Figure~\ref{fig:time_error}, the language selection module only adds a small inference cost while significantly improving the model performance. This is due to the existence of the query that is low-resource for both languages, in which switching to the opposite language may make the situation worse. In these situations, the language selection module may pick a third language to better answer the question. The full result of the performance without the language selection module can be found in Table~\ref{tab:performance_without_selection_full}.

\noindent\textbf{The Comparison between Answer Replacement and Integration.}
We further investigated the effectiveness of answer replacement and integration strategies. Given that QA setups with a golden answer may not always accommodate answer integration effectively (for example, when the answers in two different languages factually conflict), we opted for a subset in AlignBench \cite{liu2023alignbench} as our evaluation dataset. AlignBench provides a comprehensive, multi-dimensional benchmark designed to assess the alignment of LLMs in Chinese, featuring a variety of open-ended questions. To create a bilingual dataset, we translated the Chinese questions into English. For each response evaluation, we employed an LLM-as-a-judge approach, utilizing the prompt template from AlignBench. The LLM judge then assigned an overall score ranging from 1 to 10 to each LLM response. As indicated in \autoref{tab:integration}, both replacement and integration methods significantly enhanced the LLMs' performance across most datasets. Direct replacement led to more substantial improvements but also introduced a higher rate of errors, as evidenced by the performance dips in GPT-4 and Llama3-Ch. Interestingly, the integration method showed a more pronounced performance improvement in English responses, suggesting that LLMs may possess stronger capabilities for answer optimization in English than in Chinese \cite{yang2024large}.

\noindent\textbf{The Impact of Different Detection Models.} 
As we build a different low-resource detector for each language, the selection of the tokenizer and classification model would impact the training of the detector thereby influencing the overall performance. We adopt language-specific Bert and multi-lingual Bert models to train our low-resource query detector and report the result in Table~\ref{tab:det_model}. As shown in the model, using the language-specific model and tokenizer would slightly improve the result of using a multi-lingual model.

\begin{table}[t]
\centering
\small
\renewcommand\arraystretch{1.15}
\setlength{\tabcolsep}{3pt}

\begin{tabular}{c|c|c|c|c}
\toprule[1pt]
\textbf{Model}                  & \textbf{Acc.}                 & \textbf{Recall} & \textbf{Precision} & \textbf{F1.} \\
\midrule
\textbf{bert-base-chinese (ch)} & 86.64                        & 86.64           & 86.68              & 86.66       \\
\textbf{bert-uncased (en)}      & 94.98                        & 94.98           & 94.88              & 94.91       \\
\textbf{Multilingual Bert (ch)} & 86.47                        & 86.47           & 86.58              & 86.51       \\
\textbf{Multilingual Bert (en)} & {\color[HTML]{212121} 94.73} & 94.73           & 94.64              & 94.67      \\
\bottomrule[1pt]
\end{tabular}
\caption{The impact of model selection on detector training.}
\label{tab:det_model}
\end{table}

\section{Discussion on Other Approach}

As the confidence of the generated content is related to its entropy during the generation process, a natural idea is to calculate the entropy in different languages and compare them to decide which is the best language to answer the question. This approach is widely used for measuring LLMs' uncertainty and detecting hallucinations \cite{manakul2023selfcheckgpt}. However, our trial demonstrates that this approach is infeasible and achieves merely random-guess-level performance when selecting the right language for the given queries. 

To explore how to leverage entropy-related statistics to select the target language, we train a model $f$ that takes the statistics as input and outputs the selection of the language $Y$. The statistics we use for a language $l$ include the entropy of the query $E_{Q_l}$, the entropy of the response $E_{R_l}$, the perplexity of the query $P_{Q_l}$, and the perplexity of the response $P_{R_l}$. We adopt an MLP as the classification model $f:(E_{Q_l}, E_{R_l}, P_{Q_l}, P_{R_l})\rightarrow Y$ and train the model on the low-resource query dataset we construct. We trained based on SVM and random forests in Llama2-7b's output. The accuracy is no more than 60\%. This is a merely random-guess-level performance when taking the entropy-related statistics as input. We attribute this to the hallucination issue of LLMs, that the model may become over-confident even with the wrong answer \cite{groot2024overconfidence}, which indicates LLMs are not calibrated well now \cite{zhang2024calibrating}.

\vspace{-0.2cm}
\section{Conclusion}

This paper presents a method to improve the multilingual capabilities of LLMs by leveraging knowledge from various languages, which includes a low-resource knowledge detector, a process for selecting languages, and answer replacement \& integration. Our experiments show significant enhancements in performance, especially in reducing disparities across languages. Moreover, each module in our method contributes to the improvement. Overall, this study underscores the potential of LLMs to unify multilingual functions and provide insights for future research.

\newpage

\section*{Limitations}
Our method requires training a separate low-resource query detector for each language. This is not convenient as the developer of a certain language should construct a low-resource training set himself, which involves collecting language-specific data. Also, the dataset should be updated with time with the rise of the new language-specific data.

\section*{Ethics Statement}

This study adheres to ethical standards in AI research and development. We acknowledge that while our methods aim to enhance the multilingual capabilities of LLMs, they must be implemented with careful consideration of potential biases. Efforts were made to ensure that the knowledge aggregation does not disproportionately favor any particular language or cultural perspective. We also emphasize transparency in our methodologies and findings to enable scrutiny and replication by the research community. The research was conducted without utilizing any personally identifiable information, thereby safeguarding privacy and upholding data protection standards. We commit to ongoing evaluation of our methods in diverse linguistic settings to address and mitigate any emergent biases or disparities. This research seeks not only to advance technology but also to promote inclusivity and fairness in AI applications across different linguistic and cultural groups. In this paper, we utilized AI tools to aid in writing and coding, ensuring that they did not directly contribute to the writing process and that their use adheres to academic standards. Additionally, we ensured that all datasets and benchmarks used in the study comply with their intended purposes and standards.


\bibliography{custom}
\bibliographystyle{acl_natbib}

\appendix

\section{Dataset Details}
\label{app:dataset_details}

\begin{itemize}[nolistsep, leftmargin=*]
    \item \textbf{TriviaQA} \cite{joshi2017triviaqa} is a reading comprehension dataset that features more than 650,000 question-answer-evidence triples. It consists of lots of question-answer pairs created by trivia aficionados, along with independently collected evidence documents—averaging six per question—that offer robust distant supervision for responding to the queries.
    \item \textbf{HaluEval} \cite{li2023halueval} is a benchmark designed to assess how well LLMs hallucinations—unverifiable or incorrect content in their outputs. It includes a collection of generated texts and human-annotated samples that help evaluate the models' performance in detecting such errors.
    \item \textbf{HalluQA} \cite{cheng2023evaluating} is a dataset consisting of 450 carefully crafted adversarial questions that cover various domains, incorporating elements of Chinese historical culture, customs, and social phenomena. It aims to evaluate LLMs on their propensity to produce two types of errors: imitative falsehoods and factual inaccuracies.
    \item \textbf{WebQA} \cite{li2016dataset} is a large-scale, human-annotated real-world QA dataset, developed to address the scarcity of extensive real-world QA datasets for neural QA systems.
\end{itemize}

\section{Experiment Results}

\begin{figure*}
    \centering
    \includegraphics[width=1\linewidth]{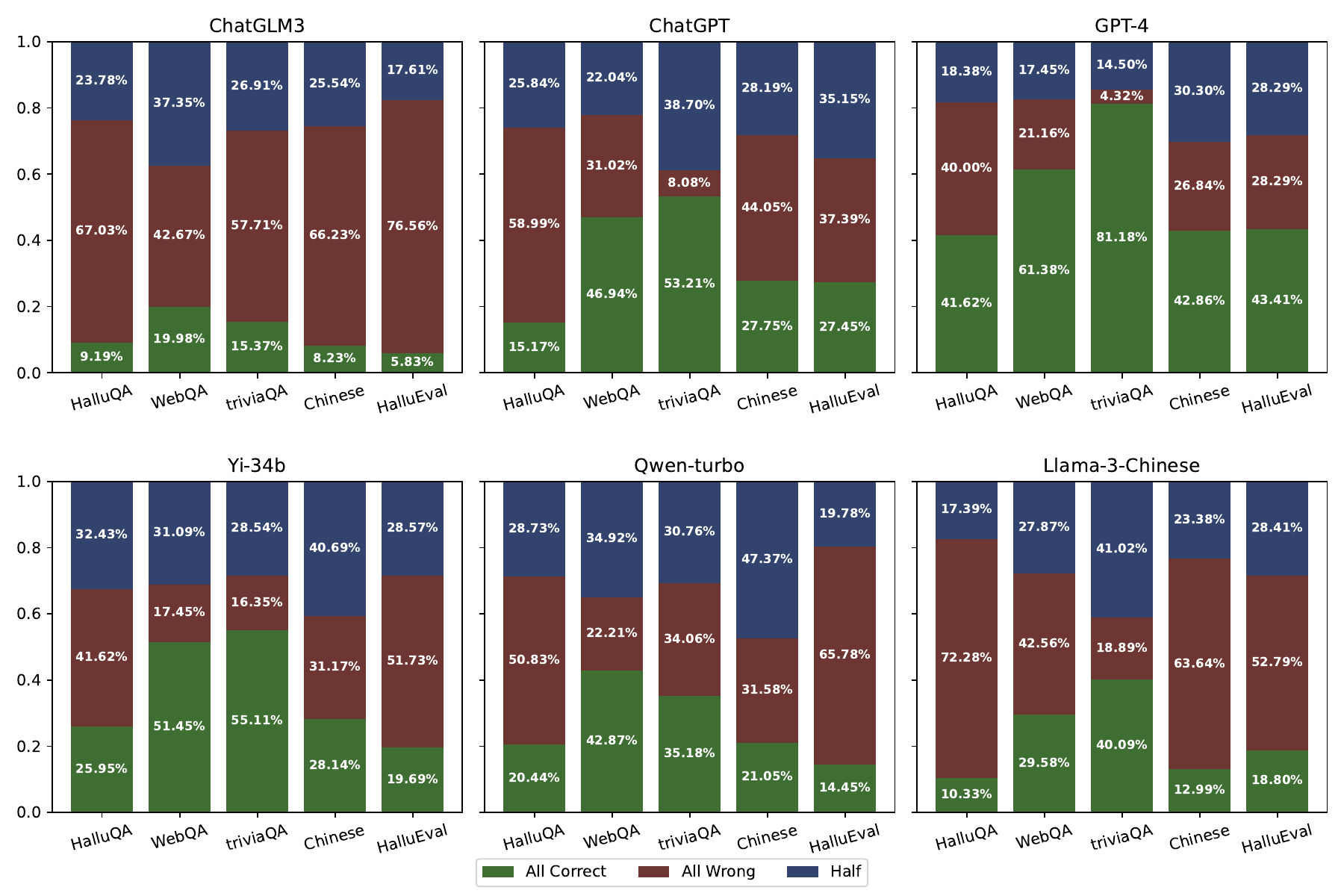}
    \caption{Performance percentage of LLMs across different datasets. `All correct' indicates that the LLMs answered correctly in both the Chinese and English datasets. `All wrong' signifies that the LLMs answered incorrectly in both datasets. `Half' denotes that the LLMs answered correctly in only one of the datasets.}
    \label{fig:motivation2}
\end{figure*}

We show the full experiment results in \autoref{tab:performance_without_detector_full}, \autoref{tab:performance_without_selection_full}, and \autoref{fig:motivation2}.

\begin{table*}[]
\centering
\small
\renewcommand\arraystretch{1.3}
\setlength{\tabcolsep}{2.2pt}

\scalebox{0.85}{\begin{tabular}{c|c|cc|cc|cc|cc|cc|cc}
\toprule[1pt]
\multicolumn{1}{c|}{}                                   &                                     & \multicolumn{2}{c|}{\textbf{ChatGLM3}}                             & \multicolumn{2}{c|}{\textbf{ChatGPT}}                              & \multicolumn{2}{c|}{\textbf{GPT-4}}                                & \multicolumn{2}{c|}{\textbf{Yi-34b}}                               & \multicolumn{2}{c|}{\textbf{Qwen-turbo}}     & \multicolumn{2}{c}{\textbf{Llama3-Ch.}}                      \\

 \cmidrule(lr){3-4}  \cmidrule(lr){5-6}  \cmidrule(lr){7-8}  \cmidrule(lr){9-10}  \cmidrule(lr){11-12} \cmidrule(lr){13-14}

\multicolumn{1}{c|}{\multirow{-2}{*}{\textbf{Dataset}}} & \multirow{-2}{*}{\textbf{Lang.}} & \textit{Origi.}           & \textit{Impro.}           & \textit{Origi.}           & \textit{Impro.}           & \textit{Origi.}           & \textit{Impro.}           & \textit{Origi.}           & \textit{Impro.}           & \textit{Origi.}           & \textit{Impro.}  & \textit{Origi.}           & \textit{Impro.}          \\
\midrule
                                                       & (en)                                & 18.03\%                         & 18.03\%                         & 57.98\%                         & 57.84\%                         & 67.13\%                         & 66.99\%                         & \cellbg[red!20]{42.86\%} & \cellbg[red!20]{41.75\%} & 29.31\%                         & 29.59\%       & 40.67\% &        40.67\%         \\
\multirow{-2}{*}{\textbf{HalluEval}}                   & (ch)                                & \cellbg[green!20]{11.23\%} & \cellbg[green!20]{17.34\%} & \cellbg[green!20]{32.07\%} & \cellbg[green!20]{52.38\%} & \cellbg[green!20]{47.99\%} & \cellbg[green!20]{65.05\%} & \cellbg[green!20]{25.10\%} & \cellbg[green!20]{39.81\%} & \cellbg[green!20]{19.35\%} & \cellbg[green!20]{26.51\%} & \cellbg[green!20]{25.35\%} & \cellbg[green!20]{37.33\%} \\
\midrule
\midrule
                                                       & {\color[HTML]{212121} (en)}         & \cellbg[green!20]{20.00\%} & \cellbg[green!20]{25.41\%} & \cellbg[red!20]{34.27\%} & \cellbg[red!20]{30.90\%} & \cellbg[green!20]{51.89\%} & \cellbg[green!20]{53.51\%} & \cellbg[green!20]{38.38\%} & \cellbg[green!20]{47.03\%} & \cellbg[green!20]{25.97\%} & \cellbg[green!20]{37.57\%} & \cellbg[red!20]{22.83\%} & \cellbg[red!20]{18.48\%} \\
\multirow{-2}{*}{\textbf{HalluQA}}                     & (ch)                                & 22.16\%                         & 22.70\%                         & \cellbg[green!20]{21.91\%} & \cellbg[green!20]{25.28\%} & \cellbg[green!20]{49.73\%} & \cellbg[green!20]{51.89\%} & 45.95\%                         & 45.95\%                         & \cellbg[red!20]{43.65\%} & \cellbg[red!20]{39.78\%} & \cellbg[green!20]{15.22\%} & \cellbg[green!20]{20.65\%}  \\
\midrule
\midrule
                                         \textbf{Chinese}              & (en)                                & \cellbg[green!20]{9.52\% } & \cellbg[green!20]{21.21\%} & 41.85\%                         & 42.73\%                         & 56.71\%                         & 57.58\%                         & \cellbg[green!20]{33.33\%} & \cellbg[green!20]{57.58\%} & \cellbg[green!20]{27.19\%} & \cellbg[green!20]{48.25\%} & \cellbg[red!20]{30.74\%} & \cellbg[red!20]{24.24\%} \\
\textbf{Domain}             & (ch)                                & \cellbg[red!20]{32.47\%} & \cellbg[red!20]{25.54\%} & 41.85\%                         & 42.29\%                         & 59.31\%                         & 58.44\%                         & \cellbg[red!20]{63.64\%} & \cellbg[red!20]{57.14\%} & 62.28\%                         & 62.28\%     & \cellbg[green!20]{18.61\% } & \cellbg[green!20]{22.51\%}                    \\
\midrule
\midrule
                                                       & (en)                                & 36.32\%                         & 36.38\%                         & 90.53\%                         & 90.31\%                         & 94.09\%                         & 93.93\%                         & 79.33\%                         & 78.90\%                         & 59.59\%                         & 59.47\%     & 77.27\% & 77.05\%                    \\
\multirow{-2}{*}{\textbf{triviaQA}}                    & (ch)                                & \cellbg[green!20]{21.33\%} & \cellbg[green!20]{32.22\%} & \cellbg[green!20]{54.60\%} & \cellbg[green!20]{83.33\%} & \cellbg[green!20]{82.77\%} & \cellbg[green!20]{92.29\%} & \cellbg[green!20]{59.43\%} & \cellbg[green!20]{76.27\%} & \cellbg[green!20]{41.53\%} & \cellbg[green!20]{53.55\%} & \cellbg[green!20]{43.92\%} & \cellbg[green!20]{66.32\%} \\
\midrule
\midrule
                                                       & (en)                                & \cellbg[green!20]{28.51\%} & \cellbg[green!20]{38.96\%} & 59.08\%                         & 58.98\%                         & \cellbg[green!20]{67.70\%} & \cellbg[green!20]{69.61\%} & \cellbg[green!20]{57.07\%} & \cellbg[green!20]{69.71\%} & \cellbg[green!20]{49.48\%} & \cellbg[green!20]{62.11\%} & \cellbg[red!20]{50.00\%} & \cellbg[red!20]{47.08\%} \\
\multirow{-2}{*}{\textbf{WebQA}}                       & (ch)                                & \cellbg[red!20]{48.69\%} & \cellbg[red!20]{42.07\%} & \cellbg[green!20]{57.35\%} & \cellbg[green!20]{59.29\%} & 72.52\%                         & 72.32\%                         & \cellbg[red!20]{76.93\%} & \cellbg[red!20]{74.12\%} & 71.12\%                         & 70.70\%   & \cellbg[green!20]{37.02\%} & \cellbg[green!20]{40.54\%}                     \\
\bottomrule[1pt]
\end{tabular}}
\caption{Six LLMs' performance on the setting without a low-resource detector.}
\label{tab:performance_without_detector_full}
\end{table*}

\begin{table*}[]
\centering
\small
\renewcommand\arraystretch{1.3}
\setlength{\tabcolsep}{2.2pt}

\scalebox{0.85}{\begin{tabular}{c|c|cc|cc|cc|cc|cc|cc}
\toprule[1pt]
\multicolumn{1}{c|}{}                                   &                                     & \multicolumn{2}{c|}{\textbf{ChatGLM3}}                             & \multicolumn{2}{c|}{\textbf{ChatGPT}}                              & \multicolumn{2}{c|}{\textbf{GPT-4}}                                & \multicolumn{2}{c|}{\textbf{Yi-34b}}                               & \multicolumn{2}{c|}{\textbf{Qwen-turbo}}     & \multicolumn{2}{c}{\textbf{Llama3-Ch.}}                      \\

 \cmidrule(lr){3-4}  \cmidrule(lr){5-6}  \cmidrule(lr){7-8}  \cmidrule(lr){9-10}  \cmidrule(lr){11-12} \cmidrule(lr){13-14}

\multicolumn{1}{c|}{\multirow{-2}{*}{\textbf{Dataset}}} & \multirow{-2}{*}{\textbf{Lang.}} & \textit{Origi.}           & \textit{Impro.}           & \textit{Origi.}           & \textit{Impro.}           & \textit{Origi.}           & \textit{Impro.}           & \textit{Origi.}           & \textit{Impro.}           & \textit{Origi.}           & \textit{Impro.}  & \textit{Origi.}           & \textit{Impro.}          \\
\midrule
                                                       & (en)                                & 18.03\%                         & 18.31\%                         & 57.98\%                         & 57.70\%                         & 67.13\%                         & 66.57\%                         & {42.86\%} & {42.72\%} & 29.31\%                         & 29.17\%   & 40.67\% & 40.25\%                       \\
\multirow{-2}{*}{\textbf{HalluEval}}                   & (ch)                                & \cellbg[green!20]{11.23\%} & \cellbg[green!20]{18.03\%} & \cellbg[green!20]{32.07\%} & \cellbg[green!20]{56.02\%} & \cellbg[green!20]{47.99\%} & \cellbg[green!20]{66.16\%} & \cellbg[green!20]{25.10\%} & \cellbg[green!20]{42.58\%} & \cellbg[green!20]{19.35\%} & \cellbg[green!20]{28.61\%} & \cellbg[green!20]{25.34\%} & \cellbg[green!20]{39.97\%} \\
\midrule
\midrule
                                                       & {\color[HTML]{212121} (en)}         & \cellbg[green!20]{20.00\%} & \cellbg[green!20]{25.95\%} & \cellbg[red!20]{34.27\%} & \cellbg[red!20]{32.02\%} & \cellbg[green!20]{51.89\%} & \cellbg[green!20]{53.51\%} & \cellbg[green!20]{38.38\%} & \cellbg[green!20]{46.49\%} & \cellbg[green!20]{25.97\%} & \cellbg[green!20]{38.67\%} & \cellbg[red!20]{22.83\%} & \cellbg[red!20]{21.74\%} \\
\multirow{-2}{*}{\textbf{HalluQA}}                     & (ch)                                & \cellbg[green!20]{22.16\%} & \cellbg[green!20]{23.78\%} & \cellbg[green!20]{21.91\%} & \cellbg[green!20]{23.60\%} & \cellbg[green!20]{49.73\%} & \cellbg[green!20]{51.35\%} & \cellbg[red!20]{45.95\%} & \cellbg[red!20]{44.86\%} & \cellbg[red!20]{43.65\%} & \cellbg[red!20]{41.44\%} & 15.21\% & 15.76\% \\
\midrule
\midrule
                                          \textbf{Chinese}             & (en)                                & \cellbg[green!20]{9.52\%}  & \cellbg[green!20]{32.03\%} & 41.85\%                         & 41.85\%                         & \cellbg[green!20]{56.71\%} & \cellbg[green!20]{59.31\%} & \cellbg[green!20]{33.33\%} & \cellbg[green!20]{61.04\%} & \cellbg[green!20]{27.19\%} & \cellbg[green!20]{58.33\%} & 
                                                       \cellbg[red!20]{30.74\%} & \cellbg[red!20]{19.48\%}\\

\textbf{Domain}              & (ch)                                & \cellbg[red!20]{32.47\%} & \cellbg[red!20]{30.74\%} & 41.85\%                         & 41.41\%                         & 59.31\%                         & 58.44\%                         & \cellbg[red!20]{63.64\%} & \cellbg[red!20]{60.17\%} & \cellbg[red!20]{62.28\%} & \cellbg[red!20]{58.33\%}  & 18.61\% & 18.61\% \\
                                                       \midrule
\midrule
                                                       & (en)                                & 36.32\%                         & 35.78\%                         & \cellbg[red!20]{90.53\%} & \cellbg[red!20]{89.09\%} & 94.09\%                         & 93.54\%                         & 79.33\%                         & 78.73\%                         & 59.59\%                         & 58.80\%      & \cellbg[red!20]{77.27\%} & \cellbg[red!20]{76.12\%}                    \\

\multirow{-2}{*}{\textbf{triviaQA}}                    & (ch)                                & \cellbg[green!20]{21.33\%} & \cellbg[green!20]{35.94\%} & \cellbg[green!20]{54.60\%} & \cellbg[green!20]{89.15\%} & \cellbg[green!20]{82.77\%} & \cellbg[green!20]{93.22\%} & \cellbg[green!20]{59.43\%} & \cellbg[green!20]{78.18\%} & \cellbg[green!20]{41.53\%} & \cellbg[green!20]{58.41\%} & \cellbg[green!20]{43.92\%} & \cellbg[green!20]{74.92\%} \\
                                                       \midrule
\midrule
                                                       & (en)                                & \cellbg[green!20]{28.51\%} & \cellbg[green!20]{44.38\%} & 59.08\%                         & 59.90\%                         & \cellbg[green!20]{67.70\%} & \cellbg[green!20]{70.81\%} & \cellbg[green!20]{57.07\%} & \cellbg[green!20]{73.72\%} & \cellbg[green!20]{49.48\%} & \cellbg[green!20]{67.70\%} &
                                                       \cellbg[red!20]{50.00\%} & \cellbg[red!20]{46.48\%}\\
\multirow{-2}{*}{\textbf{WebQA}}                       & (ch)                                & \cellbg[red!20]{48.69\%} & \cellbg[red!20]{46.99\%} & \cellbg[green!20]{57.35\%} & \cellbg[green!20]{58.88\%} & 72.52\%                         & 71.61\%                         & \cellbg[red!20]{76.93\%} & \cellbg[red!20]{74.22\%} & \cellbg[red!20]{71.12\%} & \cellbg[red!20]{69.25\%}  & \cellbg[green!20]{37.02\%} & \cellbg[green!20]{41.15\%} \\ \bottomrule[1pt]
\end{tabular}}
\caption{Six LLMs' performance on the setting without language selection.}
\label{tab:performance_without_selection_full}
\end{table*}

\section{Details of Constructed Dataset}
\label{app:constructed}

For the generated dataset, inspired by previous studies \cite{huang2024metatool, yu2023large}, we employed attribute-guided prompting to instruct LLMs to generate relevant questions on specific topics, as illustrated in \autoref{tab:topic}. We chose GPT-4 as our generation model because of its exceptional ability to follow instructions. The prompt template is shown in \autoref{fig:generation_prompt}. For the generated items, we manually evaluate the correctness of its label to ensure the data quality.

\begin{table*}[h]
\centering
\small
\renewcommand\arraystretch{1.2}
\setlength{\tabcolsep}{4pt}
\caption{Topics used for data generation.}
\label{tab:topic}
\begin{tabular}{cccc}
\toprule[1pt]
\rowcolor{gray!10!white} \multicolumn{4}{c}{\textbf{\textit{Topic Word}}} \\
\midrule
\texttt{History} & \texttt{Literature} & \texttt{Science} & \texttt{Art} \\
\midrule
\rowcolor{green!20!white} \texttt{Social Sciences} & \texttt{Technology} & \texttt{Philosophy} & \texttt{Geography} \\
\midrule
\texttt{Culture} & \texttt{Health} & \texttt{Artificial Intelligence} & \texttt{Machine Learning} \\
\midrule
\rowcolor{green!20!white} \texttt{Big Data} & \texttt{Blockchain} & \texttt{Internet of Things} & \texttt{Environmental Protection} \\
\midrule
\texttt{Sustainable Development} & \texttt{Energy} & \texttt{Finance} & \texttt{Education} \\
\midrule
\rowcolor{green!20!white} \texttt{Human Genetics} & \texttt{Artificial Life} & \texttt{Space Exploration} & \texttt{Food Science} \\
\midrule
\texttt{Sports} & \texttt{Psychology} & \texttt{Political Science} & \texttt{Economics} \\
\midrule
\rowcolor{green!20!white} \texttt{Sociology} & \texttt{Law} & \texttt{} & \texttt{} \\
\bottomrule[1pt]
\end{tabular}
\end{table*}

\begin{figure*}

\begin{tcolorbox}[
  enhanced, 
  colframe=cyan!75!black, 
  colback=white, 
  coltitle=white, 
  colbacktitle=cyan!75!black, 
  width=\linewidth, 
  arc=2mm, 
  auto outer arc, 
  boxrule=0.5pt, 
  left=10pt, 
  right=10pt, 
  drop shadow={black!50!white},
  top=10pt, 
  bottom=10pt, 
  title=\textbf{Prompt Template}, 
  fonttitle=\bfseries, 
  title code={\node[rounded corners, fill=blue!75!black, draw=none, text=white] at (frame.title) {\textbf{xxx}};}, 
  attach boxed title to top center={yshift=-2mm}, 
  boxed title style={sharp corners, size=small}, 
]
\small

Next, I will provide you with a topic, and you will assist me in generating data based on this topic. I need you to generate three categories: questions with an English background, questions with a Chinese background, and questions with no specific language.

I will provide you with some examples:

Question: Piaget believes that communication has two functions, one is the egocentric function, and the other is? \\
Category: English knowledge

Question: With one byte, how many different codes can be generated at most? \\
Category: Knowledge with no specific language

Question: What are some famous dishes from Guangdong? \\
Category: Chinese knowledge

For each type of question, you need to generate ten, a total of thirty.

You need to return the data in JSON format, as follows:
\begin{verbatim}
{
"Question": "Category",
"Question": "Category",
"Question": "Category",
"Question": "Category",
...
}
\end{verbatim}
Please generate the corresponding data in Chinese. \\
The topic I provide is: [TOPIC]

\end{tcolorbox}
\caption{Prompt template for the generated dataset.}
\label{fig:generation_prompt}
\end{figure*}

\section{Collection of Chinese Domain Dataset}
\label{app:construct_domain}
Our Chinese domain dataset consists of 227 items. This dataset encompasses knowledge and information specific to Chinese content, including aspects of geography, history, culture, and more. We sourced the content from a broad range of Chinese social media platforms and search engines. After initial curation, we conducted filtering to remove contents that cannot be accurately translated into English or may result in discrepancies in meaning upon translation, such as phrases from ancient Chinese.

\section{Answer Evaluation}

We adopt the LLM-as-a-Judge for answer evaluation in all experiments to reduce the bias that comes from keyword matching. We use GPT-4 for our evaluation due to its exceptional capability and wide application in previous studies \cite{liu2023alignbench, gao2024best}. For the five QA datasets, we use the prompt template shown in \autoref{fig:llm-as-a-judge}.

\begin{figure*}

\begin{tcolorbox}[
  enhanced, 
  colframe=cyan!75!black, 
  colback=white, 
  coltitle=white, 
  colbacktitle=cyan!75!black, 
  width=\linewidth, 
  arc=2mm, 
  auto outer arc, 
  boxrule=0.5pt, 
  left=10pt, 
  right=10pt, 
  drop shadow={black!50!white},
  top=10pt, 
  bottom=10pt, 
  title=\textbf{Prompt Template}, 
  fonttitle=\bfseries, 
  title code={\node[rounded corners, fill=blue!75!black, draw=none, text=white] at (frame.title) {\textbf{xxx}};}, 
  attach boxed title to top center={yshift=-2mm}, 
  boxed title style={sharp corners, size=small}, 
]
\small

As a helpful assistant, your task is now to help me assess the correctness of the provided answers. I will present a question along with its correct answer. Subsequently, I will also provide you with the answer you need to evaluate. If the answer to be evaluated correctly expresses the same meaning as the correct answer or contains the correct answer, then it is right. Ignore case errors. Although there are some errors in certain explanations within the answer, as long as the core answer is correct, the response is considered correct. Return me only one word: 'correct' or 'wrong'.

Here is the question and its correct answer:

Question: [QUESTION]

Answer: [ANSWER]

Here is the answer you should evaluate: [RES]

\end{tcolorbox}
\caption{Prompt template for LLM-as-a-Judge.}
\label{fig:llm-as-a-judge}
\end{figure*}

\begin{figure*}

\begin{tcolorbox}[
  enhanced, 
  colframe=cyan!75!black, 
  colback=white, 
  coltitle=white, 
  colbacktitle=cyan!75!black, 
  width=\linewidth, 
  arc=2mm, 
  auto outer arc, 
  boxrule=0.5pt, 
  left=10pt, 
  right=10pt, 
  drop shadow={black!50!white},
  top=10pt, 
  bottom=10pt, 
  title=\textbf{Prompt Template}, 
  fonttitle=\bfseries, 
  title code={\node[rounded corners, fill=blue!75!black, draw=none, text=white] at (frame.title) {\textbf{xxx}};}, 
  attach boxed title to top center={yshift=-2mm}, 
  boxed title style={sharp corners, size=small}, 
]
\small

You are a very helpful assistant. I will provide you with a question and the answers in both Chinese and English. You need to integrate the Chinese and English answers to provide the final answer. During the integration process, you need to follow these rules:

1. You should primarily refer to the Chinese answer, appropriately integrating parts of the English answer.

2. If there is a factual conflict between the English and Chinese answers, you must refer to the Chinese answer.

3. The integrated answer should be of higher quality than the individual answers and better address the corresponding question.

4. The integrated answer must be all in English

Question: [[Q]]

Chinese answer: [[CH\_RES]]

English answer: [[EN\_RES]]

\end{tcolorbox}
\caption{Prompt template for integration (For the situation when the selected language is English).}
\label{fig:integration_en}
\end{figure*}

\begin{figure*}

\begin{tcolorbox}[
  enhanced, 
  colframe=cyan!75!black, 
  colback=white, 
  coltitle=white, 
  colbacktitle=cyan!75!black, 
  width=\linewidth, 
  arc=2mm, 
  auto outer arc, 
  boxrule=0.5pt, 
  left=10pt, 
  right=10pt, 
  drop shadow={black!50!white},
  top=10pt, 
  bottom=10pt, 
  title=\textbf{Prompt Template}, 
  fonttitle=\bfseries, 
  title code={\node[rounded corners, fill=blue!75!black, draw=none, text=white] at (frame.title) {\textbf{xxx}};}, 
  attach boxed title to top center={yshift=-2mm}, 
  boxed title style={sharp corners, size=small}, 
]
\small

你是一个非常有帮助的助手。我将给你提供一个问题，以及该问题的中英文的答案。你需要融合中英文答案，给出最终的答案。在融合答案的过程中，你需要遵循下面的规则：

1. 你需要着重参考英文的答案，适当融合部分中文的答案。

2. 如果英文的答案与中文的答案发生事实性冲突，你必须参考英文的答案。

3. 融合后的答案应该比融合前的答案具有更高的质量，更好地回答对应的问题。

4. 融合后的答案必须全都是中文。

问题：[[Q]]

中文答案：[[CH\_RES]]

英文答案：[[EN\_RES]]

\end{tcolorbox}
\caption{Prompt template for integration (For the situation when the selected language is Chinese).}
\label{fig:integration_ch}
\end{figure*}

\section{Prompt Template}
\label{app:prompt_template}

We show the prompt template used in our study in \autoref{fig:generation_prompt}, \autoref{fig:llm-as-a-judge}, \autoref{fig:integration_ch}, \autoref{fig:integration_en}, \autoref{fig:selection_ch}, and \autoref{fig:selection_en}.

\section{Screenshots of Human Evaluation}

\begin{figure*}
    \centering
    \includegraphics[width=1\linewidth]{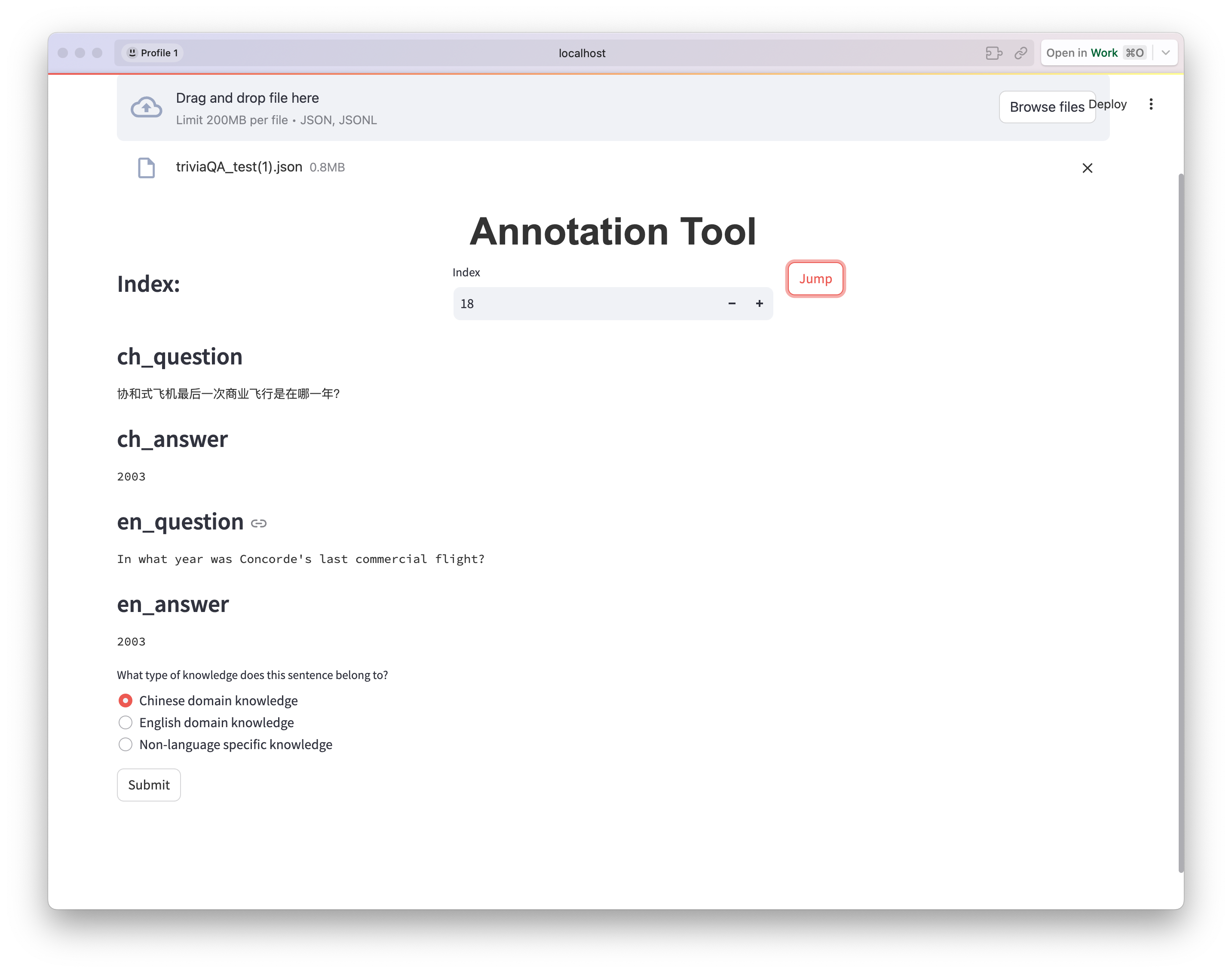}
    \caption{Screenshot of human annotation (1).}
    \label{fig:screen1}
\end{figure*}

\begin{figure*}
    \centering
    \includegraphics[width=1\linewidth]{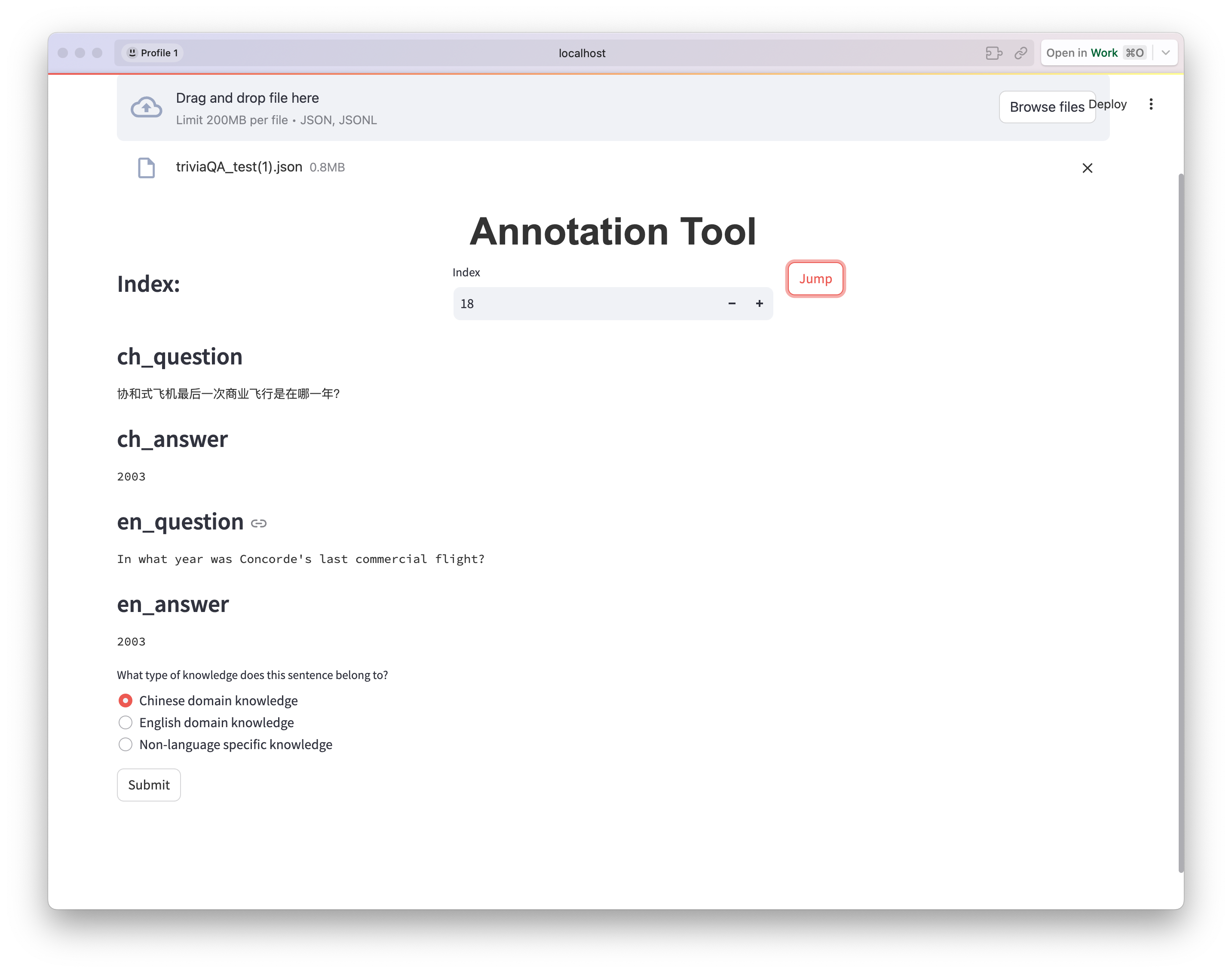}
    \caption{Screenshot of human annotation (2).}
    \label{fig:screen2}
\end{figure*}

We show the screenshots of human evaluation in \autoref{fig:screen1} and \autoref{fig:screen2}.

\begin{figure*}
\begin{tcolorbox}[
  enhanced, 
  colframe=cyan!75!black, 
  colback=white, 
  coltitle=white, 
  colbacktitle=cyan!75!black, 
  width=\linewidth, 
  arc=2mm, 
  auto outer arc, 
  boxrule=0.5pt, 
  left=10pt, 
  right=10pt, 
  drop shadow={black!50!white},
  top=10pt, 
  bottom=10pt, 
  title=\textbf{Prompt Template}, 
  fonttitle=\bfseries, 
  title code={\node[rounded corners, fill=blue!75!black, draw=none, text=white] at (frame.title) {\textbf{xxx}};}, 
  attach boxed title to top center={yshift=-2mm}, 
  boxed title style={sharp corners, size=small}, 
]
\small

As a helpful assistant, you need to categorize an English question, considering that the background of this question is not common in an English environment. Therefore, you need to choose the most suitable language for this question. You need to analyze the required language context for the question first, and then tell me at the end which language you think is most suitable to answer the question. The question is as follows: 

\end{tcolorbox}
\caption{Prompt template for language selection (For the query in English).}
\label{fig:selection_en}
\end{figure*}

\begin{figure*}
\begin{tcolorbox}[
  enhanced, 
  colframe=cyan!75!black, 
  colback=white, 
  coltitle=white, 
  colbacktitle=cyan!75!black, 
  width=\linewidth, 
  arc=2mm, 
  auto outer arc, 
  boxrule=0.5pt, 
  left=10pt, 
  right=10pt, 
  drop shadow={black!50!white},
  top=10pt, 
  bottom=10pt, 
  title=\textbf{Prompt Template}, 
  fonttitle=\bfseries, 
  title code={\node[rounded corners, fill=blue!75!black, draw=none, text=white] at (frame.title) {\textbf{xxx}};}, 
  attach boxed title to top center={yshift=-2mm}, 
  boxed title style={sharp corners, size=small}, 
]
\small

作为乐于助人的助理，您需要将一个中文问题进行分类，考虑到该问题背景在中文环境中并不常见，因此您需要返回最适合该问题的语言。你需要首先对问题所需要的语言环境进行分析，然后在最后告诉我你返回的最适合回答该问题的语言。问题如下：

\end{tcolorbox}
\caption{Prompt template for language selection (For the query in Chinese).}
\label{fig:selection_ch}
\end{figure*}

\end{CJK}

\end{document}